%% file: 00_arxiv.tex
\documentclass{article}
\PassOptionsToPackage{dvipsnames}{xcolor}

\usepackage{microtype}
\usepackage{graphicx}
\usepackage{subfigure}
\usepackage{booktabs} 
\usepackage{pifont}
\usepackage{hyperref}

\usepackage[accepted]{arxiv2025}

\usepackage{amsmath}
\usepackage{amssymb}
\usepackage{mathtools}
\usepackage{amsthm}

\usepackage[capitalize,noabbrev]{cleveref}

\usepackage[textsize=tiny]{todonotes}

\usepackage{bbm}
\usepackage{pifont}
\usepackage{makecell}
\usepackage{longtable}
\usepackage{array}
\usepackage{multirow}
\usepackage[most]{tcolorbox}
\definecolor{customblue}{RGB}{192,233,255}
\definecolor{customgold}{RGB}{255, 215, 0}
\definecolor{customgreen}{RGB}{154,235,138}
\definecolor{customred}{RGB}{230,124,130}

\input{math_commands}

\newcommand{\cmark}{\ding{51}}%
\newcommand{\xmark}{\ding{55}}

\icmltitlerunning{Position: Model Collapse Does Not Mean What You Think}

\begin{document}

\twocolumn[
\icmltitle{Position: Model Collapse Does Not Mean What You Think}



\icmlsetsymbol{equal}{*}

\begin{icmlauthorlist}
\icmlauthor{Rylan Schaeffer}{stanfordcs}
\icmlauthor{Joshua Kazdan}{stanfordstats}
\icmlauthor{Alvan Caleb Arulandu}{harvard}
\icmlauthor{Sanmi Koyejo}{stanfordcs}
\end{icmlauthorlist}

\icmlaffiliation{stanfordcs}{Stanford Computer Science}
\icmlaffiliation{stanfordstats}{Stanford Statistics}
\icmlaffiliation{harvard}{Harvard University}

\icmlcorrespondingauthor{Rylan Schaeffer}{rschaef@cs.stanford.edu}
\icmlcorrespondingauthor{Sanmi Koyejo}{sanmi@cs.stanford.edu}

\icmlkeywords{Machine Learning, ICML, model collapse, model-data feedback loops, language models, statistics, generative modeling, stochastic processes}

\vskip 0.3in
]



\printAffiliationsAndNotice{}  

\begin{abstract}
The proliferation of AI-generated content online has fueled concerns over \textit{model collapse}, a degradation in future generative models' performance when trained on synthetic data generated by earlier models. Industry leaders, premier research journals and popular science publications alike have prophesied catastrophic societal consequences stemming from model collapse.
In this position piece, we contend this widespread narrative fundamentally misunderstands the scientific evidence.
We highlight that research on model collapse actually encompasses eight distinct and at times conflicting definitions of model collapse, and argue that inconsistent terminology within and between papers has hindered building a comprehensive understanding of model collapse.
To assess how significantly different interpretations of model collapse threaten future generative models, we posit what we believe are realistic conditions for studying model collapse and then conduct a rigorous assessment of the literature's methodologies through this lens.
While we leave room for reasonable disagreement, our analysis of research studies, weighted by how faithfully each study matches real-world conditions, leads us to conclude that certain predicted claims of model collapse rely on assumptions and conditions that poorly match real-world conditions, and in fact several prominent collapse scenarios are readily avoidable.
Altogether, this position paper argues that model collapse has been warped from a nuanced multifaceted consideration into an oversimplified threat, and that the evidence suggests specific harms more likely under society's current trajectory have received disproportionately less attention.
\end{abstract}

\input{01_introduction}

\input{02_definitions}
\input{03_realistic_conditions}
\input{04_assessing_results}
\input{05_position}

\clearpage

\nocite{kazdan2024accumulating}
\bibliography{references}
\bibliographystyle{icml2025}

\clearpage
\input{99_appendix}

\end{document}

%% file: math_commands.tex
\usepackage{amsmath,amsfonts,bm,amssymb,amsthm}

\DeclareMathOperator{\defeq}{\stackrel{\text{def}}{\;=\;}}

\def\eqref#1{equation~\ref{#1}}

\def\1{\bm{1}}

\DeclareMathAlphabet{\mathsfit}{\encodingdefault}{\sfdefault}{m}{sl}
\SetMathAlphabet{\mathsfit}{bold}{\encodingdefault}{\sfdefault}{bx}{n}



%% file: 01_introduction.tex
\section{Introduction}
\label{sec:introduction}

The rapid surge of AI-generated content has sparked intense debate about potential ramifications of training future generative AI models on datasets containing synthetic data generated by previous models.
One especially concerning prediction is \textit{model collapse}: a phenomenon whereby future generative models fail due to being trained on synthetic data.
Model collapse has captured attention at the highest levels of academia and industry:
\textit{Nature} prominently featured model collapse in 2024 \citep{gibney2024nature} alongside accompanying research suggesting that AI models trained on synthetic data would suffer catastrophic degradation in performance \citep{shumailov2024curseofrecursion}, while prominent science and news outlets like \textit{Scientific American} and the \textit{Wall Street Journal} amplified these concerns, writing ``a training diet of AI-generated text, even in small quantities, eventually becomes poisonous to the model being trained." \citep{rao2023aipoison} and that ``feeding a model text that is itself generated by AI is considered the computer-science version of inbreeding" \citep{seetharaman2024data}. Meanwhile, some industry leaders have highlighted model collapse as a critical challenge for the future of AI development and deployment \citep{wang2024tweet}.

\textbf{In this position piece, we argue that this widespread narrative of model collapse, which describes a bleak future filled with polluted pretraining data and useless generative models, oversimplifies or misinterprets both the precise scientific claims and their underlying assumptions and mechanisms.} Through careful analysis, we identify three critical gaps between the prevailing discourse and research reality:

\begin{figure*}
    \centering
    \includegraphics[width=\linewidth]{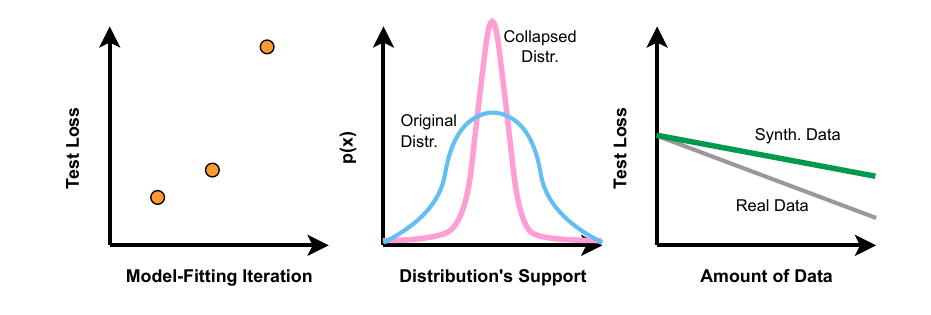}
    \caption{\textbf{Model Collapse Has Been Defined in Multiple and Sometimes Conflicting Ways.} By hand-annotating 28 prior research publications, we identify 8 definitions of model collapse (Sec.~\ref{sec:multiple_definitions}). The 8 definitions can be loosely grouped into three families: (1) the behavior of the test loss on real data over model-fitting iterations (left), (2) the deformation of the real data distribution over model-fitting iterations (center) and (3) the scaling behavior of the test loss with respect to typical scaling quantities such as the amount of data (right).
    }
    \label{fig:schematic}
\end{figure*}

First, we reveal that the term ``model collapse" encompasses eight definitions of performance degradation in model-data feedback loops. This multiplicity of definitions, used inconsistently between papers and at times inconsistently within papers, has resulted in papers talking past one another, thereby hindering development of a comprehensive understanding of likely futures for frontier deep generative models. We argue specific failure modes should be explicitly identified and discussed alongside comparable results.

Second, we posit trends that we believe faithfully describe common practices of leading AI labs pretraining frontier AI systems on web-scale data: increasing compute, improving data quality, and expanding datasets of real and synthetic data. One key point that we emphasize here is that many prominent model collapse papers assume data are entirely deleted after each model-fitting iteration and that subsequent models are trained entirely on synthetic data generated by their predecessors, which we argue is not realistic.

Third, we weigh different results in the model collapse literature based on the plausibility of their assumptions along multiple axes of consideration to assess which notions of model collapse pose significant and likely threats to future frontier AI models. We argue that some definitions of model collapse do not correspond to catastrophic outcomes under our realistic assumptions, and most concerning collapse predictions emerge from implausible experimental setups. However, there are very real threats to tails of the data distribution that should be taken seriously.

Altogether we argue that model collapse has been inflated from a precise and important technical consideration into a mischaracterized and overstated threat.
While synthetic data poses genuine challenges that warrant careful study, our analysis reveals that the most widely stated collapse scenarios can be avoided through standard ongoing practices in model development and dataset curation. Instead of worrying about unrealistic catastrophic notions of model collapse, by adopting a more realistic perspective, we can re-orient to focus on the real issues of diversity collapse happening now \citep{zhang2024forcing, padmakumar2024doeswritinglanguagemodels,murthy2024fishfishseaalignment,wu2024generativemonoculturelargelanguage}.
This position paper aims to clarify the scientific discourse around model collapse, propose best practices for future work on the subject, and redirect research attention towards understanding how to generate and curate synthetic data that improves future frontier AI systems while mitigating failure modes.

%% file: 02_definitions.tex
\section{Definitions of Model Collapse}
\label{sec:multiple_definitions}

\begin{figure*}[t!]
    \centering
    \includegraphics[width=\linewidth]{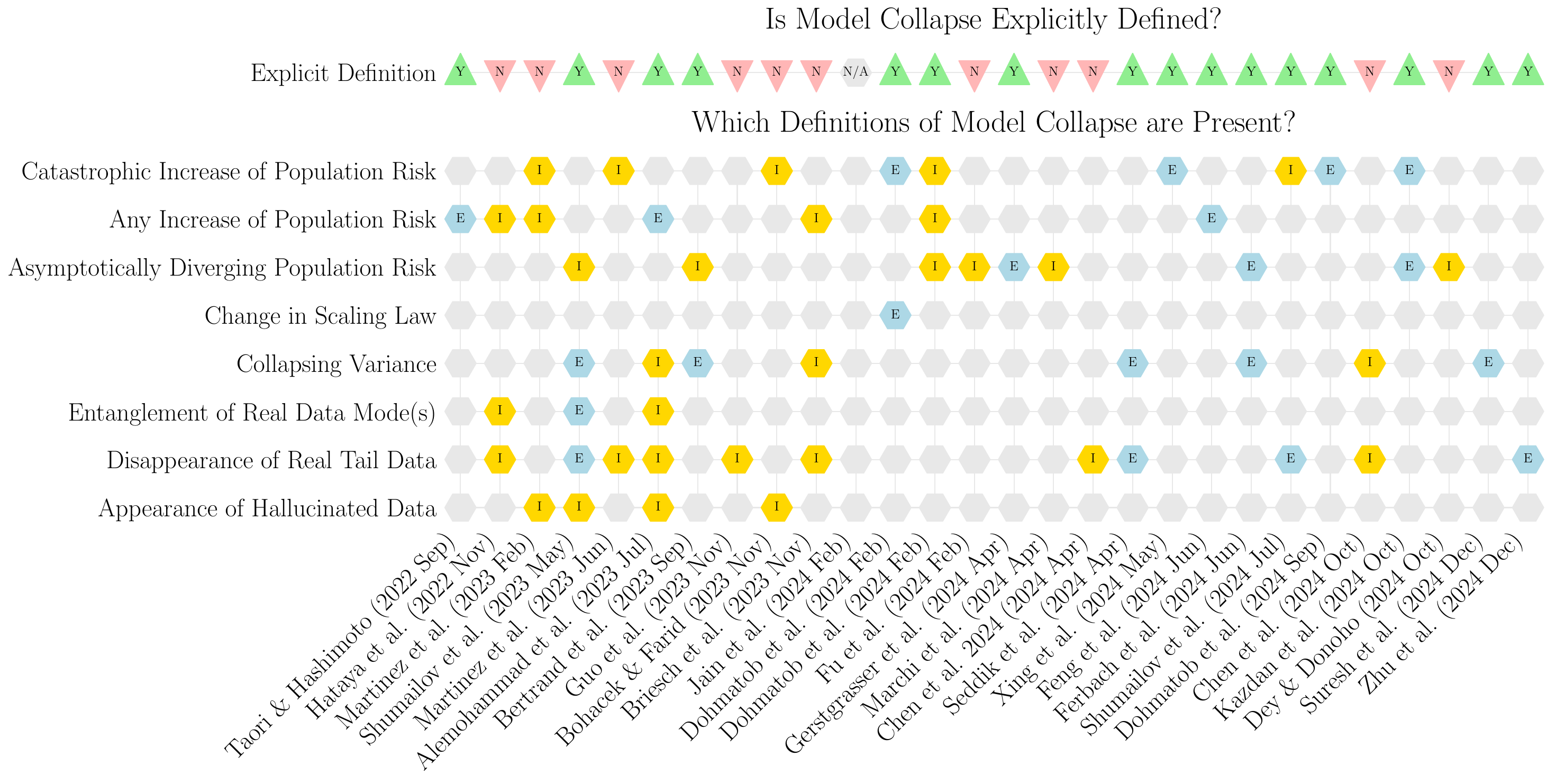}
    \caption{\textbf{Model Collapse has been defined in multiple and sometimes conflicting ways.} We conduct a meta-analysis of research papers on model collapse. Top: We identify which papers offer \textit{any} explicit definition of model collapse (\textbf{Yes \textcolor{customgreen}{(Y)}} or \textbf{No \textcolor{customred}{(N)}}), broadly construed. Bottom: We identify which definition(s) of model collapse each paper uses for its experimental and/or mathematical results, either \textbf{explicitly \textcolor{customblue}{(E)}} or \textbf{implicitly \textcolor{customgold}{(I)}}. Our annotations reveal that research on model collapse is based on multiple definitions that we will show sometimes conflict between papers and even within individual papers.
    }
    \label{fig:definitions_meta_analysis}
\end{figure*}

We begin with a non-obvious but critical point: the model collapse literature has at least eight different definitions based on different notions of model performance degradation.
As evidence, we hand-annotated twenty-eight prominent prior research publications on model collapse to determine which papers offer explicit definition(s) of model collapse (Fig.~\ref{fig:definitions_meta_analysis}), where we define explicit as either (1) any mathematical definition, or (2) any precise verbal description of the failure behavior \textit{independent} from results. We additionally categorize which definition(s) of model collapse each paper uses, perhaps implicitly, across any and all mathematical and empirical results (Fig.~\ref{fig:definitions_meta_analysis} bottom). 
We find that many papers do not offer explicit definitions of model collapse and also sometimes use multiple definitions, leading to a lack of specificity and apparent contradictions both within single works and across multiple works.

Before delving into the eight definitions, we first define some shared terminology.
In general, we consider model-data feedback loops whereby $f_t$ is the $t$-th generative model, and we study the behavior of a sequence of generative models $(f_t)_t$ that are iteratively fit to data and then sampled from. We call data sampled from a generative model \textit{synthetic data}. We can evaluate the quality of a generative model in multiple ways. One prominent way is via the \textit{population risk}, defined as the expected loss over the entire real data distribution $\mathbb{E}_{x \sim P}[\ell(f_t(x))]$, where $x$ is some real datum, $P$ is the real data distribution, and $\ell$ is the loss function \citep{vapnik1991principles}.
Another way to evaluate the quality of a generative model is by its tail risk, defined as the expected loss conditioned on tail events $\mathbb{E}_{x \in \text{Tail}(P)}[\ell(f_t(x))]$, where $\text{Tail}(P)$ informally represents real data with low probability.
Population risk provides a holistic view of performance, but may mask specific failure modes \citep{mark2024estimating,kozerawski2022taming}, whereas tail risk can reveal degradations in edge cases even when population risk remains stable \citep{hoffmann2020tail, mark2024estimating}.
There are many other salient properties of generative models one can use to assess whether the models collapse.

\begin{enumerate}

    \item \label{def:catastrophic} \textbf{Catastrophic Increase of Population Risk} \citep{dohmatob2024strong, bertrand2023stability, kazdan2024collapsethriveperilspromises}: Perhaps the most colloquial definition, model collapse is a critical and rapid degradation in model performance due to the presence of synthetic data, as measured by population risk. We note that what constitutes catastrophic is often undefined.

    \item \label{def:eps} \textbf{\text{Any} Increase of Population Risk} \citep{alemohammad2023self, dohmatob2024strong}: Under this strict definition, model collapse occurs if there is \textit{any} increase in population risk when training with synthetic data compared to training with real data alone.

    \item \label{def:diverge} \textbf{Asymptotically Diverging Population Risk} \citep{gerstgrasser2024model, kazdan2024collapsethriveperilspromises, dey2024universality}: This definition considers model collapse to occur when the population risk grows without bound over successive model-fitting iterations. This represents a fundamentally unstable learning dynamic where each iteration of synthetic data generation and training leads to progressively worse performance.

    \item \label{def:var} \textbf{Collapsing Variance} \citep{alemohammad2023self, shumailov2023curse, bertrand2023stability} Model collapse here is when variance (or diversity) trends towards $0$ and the learned distributions tend towards delta-like functions over successive model-fitting iterations.

    \item \label{def:scaling} \textbf{Change in Scaling Law} \citep{dohmatob2024tale}: In this view, model collapse occurs if the governing scaling behavior changes due to the presence of synthetic data. Specifically, model collapse occurs if the relationship between model performance and training data size deviates from the expected scaling behavior observed with real data.

    \item \label{def:modes} \textbf{Disappearance of or Entanglement of Real Data Mode(s)} \citep{alemohammad2023self}: Sometimes called ``Mode Collapse" \citep{goodfellow2014gans,lucic2018gans,brock2019largescalegantraining}, model collapse here is defined by the presence of synthetic data preventing the model from learning particular modes of the real data distribution or causing the model to blur different data modes together.

    \item \label{def:tails} \textbf{Disappearance of Real Tail Data} \citep{shumailov2023curse, wyllie2024fairnessfeedbackloopstraining,shumailov2024curseofrecursion}: Sometimes called ``coverage collapse'' \citep{zhu2024synthesizetextdatamodel}, model collapse here occurs when  synthetic data leads to the under-representation of data from the tail of the distribution, leading to models that can only handle common cases but fail on rare ones. The disappearance of real tail data can be more subtle and more narrow than the generative model losing all diversity (Def.~\ref{def:var}).

    \item \label{def:hallucinate} \textbf{Appearance of Hallucinated Data} \citep{shumailov2023curse, alemohammad2023self,bohacek2023nepotistically}: Model collapse occurs when the sequence of models begin producing fully-synthetic data not supported by the original real data's distribution.
\end{enumerate}

We note that the definitions can themselves be loosely clustered into three families (Fig.~\ref{fig:schematic}): (i) population risk degrading (Definitions~\ref{def:catastrophic}, \ref{def:eps} and \ref{def:diverge}), (ii) distributions deforming from their original shape (Definitions~\ref{def:var},~\ref{def:scaling},~\ref{def:modes},~\ref{def:tails}, and \ref{def:hallucinate}), and (iii) decreasing value from additional data (Definition~\ref{def:scaling}). 

\subsection{Intra-Paper Definitions Can Cause Confusion}

The differences between different definitions of model collapse can be slippery, and understandably, authors sometimes move between them in the course of a paper. However, model collapse is a technical phenomenon that requires definitional rigor to properly characterize.
In this section, we use a prominent prior work to demonstrate how easy it can be to slip between definitions. Our intention is not to call out this specific work, but rather demonstrate how a seemingly reasonable treatment of model collapse definitions can have serious implications for interpreting results.

We consider \citet{shumailov2023curse}, which admirably provides explicit definitions of two different types of model collapse: ``We separate two special cases: early model collapse and late model collapse. In early \textit{model collapse} the model begins losing information about the tails of the distribution; in the late \textit{model collapse} the model entangles different modes of the original distributions and converges to a distribution that carries little resemblance to the original one." These correspond to our Definitions~\ref{def:tails} and \ref{def:modes}, respectively.

However, the paper presents results on model collapse that fall under different definitions. Firstly, \citet{shumailov2023curse} demonstrate Definition~\ref{def:diverge} in their Sections 4.2 and 4.3. We lightly generalize their results for clarity and generality. We consider repeatedly fitting multivariate Gaussians to data and sampling from the fitted Gaussians. We begin with $n$ \textit{real} data drawn from a multivariate Gaussian with mean $\mu^{(0)}$ and covariance $\Sigma^{(0)}$:
\begin{equation*}
    X_1^{(0)}, ..., X_n^{(0)} \sim_{i.i.d.} \mathcal{N}(\mu^{(0)}, \Sigma^{(0)}).
\end{equation*}
For model fitting, we compute the unbiased mean and covariance of the most recent data:
\begin{align*}
    \hat{\mu}^{(t+1)} &\defeq \frac{1}{n} \sum_{j=1}^n X_j^{(t)}\\
    \hat{\Sigma}^{(t+1)} &\defeq \frac{1}{n - 1} \sum_{j=1}^n (X_j^{(t)} - \hat{\mu}^{(t+1)}) (X_j^{(t)} - \hat{\mu}^{(t+1)})^T
\end{align*}
and then draw $n$ new synthetic samples from a Gaussian with the most recently fit parameters.
In this model-data feedback loop, \citet{shumailov2023curse} proved that as the model-fitting iteration $t \rightarrow \infty$, the population risk as measured by the expected squared Wasserstein distance between the most recent multivariate Gaussian and the original multivariate Gaussian diverges asymptotically:
\begin{equation*}
     \mathbb{E}[\mathbb{W}_2^2(\mathcal{N}(\hat{\mu}^{(t)}, \hat{\Sigma}^{(t)}) \;, \; \mathcal{N}(\mu^{(0)}, \Sigma^{(0)}))] \rightarrow \infty.
\end{equation*}
This result demonstrates that the population risk diverges asymptotically, which aligns with neither of the two definitions of model collapse stated at the outset of the paper.
Is it possible that the population risk is diverging \textit{because} of one of the two other definitions?
Recall that the squared Wasserstein distance between two Gaussians has two terms: a contribution from the means and a contribution from the covariances:
\begin{align*}
    &\mathbb{E}[\mathbb{W}_2^2(\mathcal{N}(\hat{\mu}^{(t)}, \hat{\Sigma}^{(t)}), \mathcal{N}(\mu^{(0)}, \Sigma^{(0)}))] = \\
    &\underbrace{||\hat{\mu}^{(t)} - \mu^{(0)}||_2^2}_{\rightarrow \; \infty}  \quad + \\
    &\underbrace{\operatorname{Tr} \big( \hat{\Sigma}^{(t)} + \Sigma^{(0)} - 2 ((\Sigma^{(0)})^{1/2}  \hat{\Sigma}^{(t)} (\Sigma^{(0)})^{1/2})^{1/2} \big)}_{\rightarrow \; 0}.
\end{align*}
Thus, while the variance collapses and tails do vanish, neither is the cause of the population risk diverging.
Rather, the population risk diverges because the sequence of means $(\hat{\mu}^{(t)})_t$ randomly walks away from the ground truth mean $\mu^{(0)}$.
This is one example in which a casual reader might fail to notice that while the population risk \textit{is} indeed asymptotically diverging, such undesirable behavior is attributable to neither real data tails disappearing quickly nor to real data modes entangling slowly.

The same paper also demonstrates how definitions of model collapse can go beyond subtle confusion to explicit contradiction. \citet{shumailov2023curse} experimentally demonstrate model collapse under a \textit{fourth} definition of model collapse: in their Section 5.2, the authors consider finetuning sequences of \texttt{OPT-125M} language models \citep{zhang2022optopenpretrainedtransformer} initially on \texttt{wikitext2} \citep{merity2016pointer} and then subsequently on the models' own generated outputs. The authors' Figure 10 shows that while the population risk (measured by test perplexity) initially increases, it then decreases and converges to a plateau about $12.5\% - 50\%$ above the population risk of the first model.
Indeed, the authors find that ``Over the generations models tend to produce samples that the original model trained with real data is more likely to produce."
We believe that under Definitions~\ref{def:diverge}, \ref{def:modes}, and \ref{def:tails}, these results suggest that model collapse has not occurred.
However, the authors label this result as model collapse because later generations trained on purely synthetic data begin introducing fully synthetic tail data over time: ``later generations start producing samples that would never be produced by the original model, i.e., they start misperceiving reality based on errors introduced by their ancestors."
Thus, this appearance of hallucinated data qualifies as model collapse under Definition~\ref{def:hallucinate}.


When a paper shifts definitions without explicitly acknowledging the change, it creates a cascade of problems undermining scientific clarity. Readers interpret results through the lens of the initially stated definitions, creating a false sense that all discussed phenomena represent the same underlying issue when they may be fundamentally distinct. This leads to misattribution of causes and effects, as demonstrated in the Gaussian example where population risk divergence was incorrectly associated with tail data disappearance rather than mean drift. Such definitional inconsistency fosters overgeneralization of results, hampering cross-study comparisons and potentially prompting inappropriate technical responses or policy decisions. Most critically, when technical concepts like model collapse require precise characterization, unstated definitional shifts prevent the formation of a stable framework for interpreting claims, ultimately contributing to broader confusion about which phenomena deserve concern and how they might be addressed.

\subsection{Inter-Paper Definitions Can Cause Confusion}

To demonstrate how different researchers can look at the same results and reach different conclusions regarding model collapse, \citet{alemohammad2023self} studied a \texttt{FFHQ-StyleGAN2} \citep{karras2019stylegan2} trained on synthetic data and found that the population risk as measured by Frechet Inception Distance (FID) \citep{heusel2018ganstrainedtimescaleupdate} increased $2\times$ by the 5th model-fitting iteration and then plateaued. The authors declared this result constituted model collapse because the authors had implicitly defined model collapse as \textit{any} increase in the population risk (Definition ~\ref{def:eps}).
\citet{gerstgrasser2024model} then questioned this claim that the models had collapsed, writing, ``Figure 7 from \citet{alemohammad2023self} shows that linearly accumulating data (“Synthetic augmentation loop”) causes poor behavior to plateau with the number of model-fitting
iterations [...] We believe is that our evidence and their evidence is more consistent with the conclusion that accumulating data avoids model collapse and does not merely delay it."
This apparent disagreement was because \citet{gerstgrasser2024model} had defined model collapse as asymptotically diverging population risk (Definition ~\ref{def:diverge}).
Thus, while looking at the exact same figure, the researchers came to differing conclusions because they were operating under different definitions.

\subsection{Stating and Adhering to Definitions Improves Clarity and Drives Progress}

In this position paper, our intention is not to argue in favor of specific definitions of model collapse or call out authors whose definitions we disagree with.
Rather, we hope to emphasize that model collapse is a multifaceted phenomenon: under the same results, model collapse can simultaneously ``occur" and ``not occur" in different researchers' opinions; in Appendix~\ref{app:sec:case_study}, we include a case study of how confusing and entangled scientific insights can become on account of different definitions and methodologies.
This makes building a comprehensive understanding of model collapse difficult, which can be especially concerning to specific communities.
For instance, search providers like Google, Bing, and Perplexity may pay an especially high penalty if their models are trained on hallucinated facts, whereas the disappearance of real tail data can disproportionately affect marginalized groups or historically disadvantaged communities \citep{blodgett2016demographic,bender2018data,noble2018algorithms,shah2020predictive,jo2020lessons,koenecke2020racial,bender2021dangers,hutchinson2021towards,ji2023survey}. 
By clarifying different definitions of model collapse and adhering to those definitions, researchers can build a better understanding of model collapse with greater nuance for what causes different failures modes and what actions can be taken to prevent each. Together, these definitions can form a ``model collapse profile" which can characterize the different ways in which models worsen over time. Research on the harms of synthetic data should explicitly note the relevant aspects of the model collapse profile they study.

%% file: 03_realistic_conditions.tex
\section{Realistic Conditions for Studying Model Collapse}
\label{sec:realistic_assumptions}

Our goal is to understand likely outcomes as humanity pre-trains future frontier AI systems on web-scale datasets containing a mixture of real and synthetic data. We focus on pre-training both because the literature has (\citet{shumailov2023curse}'s ``What will happen to GPT-{n} once LLMs contribute much of the language found online?") and because pre-training's tremendous capital and operating expenses render missteps extremely costly.
Motivated by this goal, we posit trends that we believe describe the current trajectory of pre-training practices of frontier AI systems:
\begin{enumerate}
    \item \textbf{Increasing Pre-training Compute}: The total floating point operations used for pre-training has been rapidly increasing. For example, Meta pre-trained Llama 1 using 2k GPUs, Llama 2 using 4k GPUs and Llama 3 using 16k GPUs \citep{goyal2024llamatweet}, and OpenAI recently announced a \$500B initiative to increase compute capacity of the United States \citep{openai2025stargate}, a fraction of which will be allocated for pre-training.
    \item \textbf{Increasing Pre-training Data}: The amount of data used has been rapidly increasing. For example, Meta's series of Llama language models were pre-trained on increasing amounts of data: 1.4 trillion tokens for Llama 1, 2 trillion tokens for Llama 2, and most recently, 15 trillion tokens for Llama 3. While pre-training data will inevitably max out, recent estimates place the total number of available language pre-training tokens at 1 quadrillion (1000 trillion) tokens \citep{villalobos2024rundatalimitsllm}.
    \item \textbf{Increasing Quality of Pre-training Data}: Over time, pre-training data is becoming increasingly higher quality on account of  pre-training data teams developing better filtering techniques to ensure high-quality training data \citep{gao2020pile800gbdatasetdiverse,penedo2024finewebdatasetsdecantingweb,li2024datacomplmsearchgenerationtraining}. Moreover, whatever synthetic data is shared online is increasingly higher quality since models are improving over time \cite{kiela2021dynabench,kiela2023plottingprogress,maslej2024stanfordhaiaiindex}.
    \item \textbf{Synthetic Data Accumulating Alongside Real Data}: Real data are not deleted en masse after each iteration of model pre-training. When synthetic data are generated and released online, they amasses alongside prior data and new real data.
    \item \textbf{Decreasing Proportion of Real Data}: The fraction of real data relative to total (real plus synthetic) data is decreasing over time.  However, whether the fraction of real data will asymptote to zero is unclear, a point we will return to later.
\end{enumerate}

\begin{figure*}[t!]
    \includegraphics[width=0.8\linewidth]{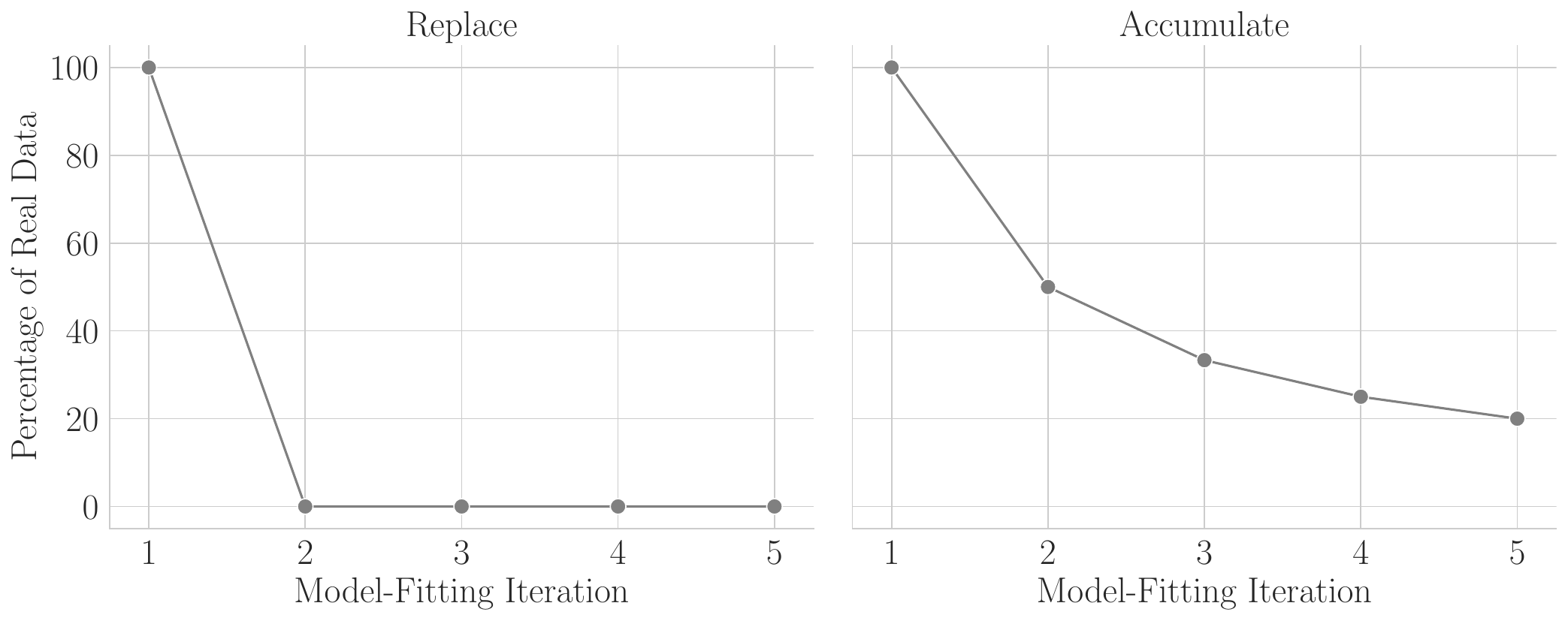}\\
    \includegraphics[width=0.97\linewidth]{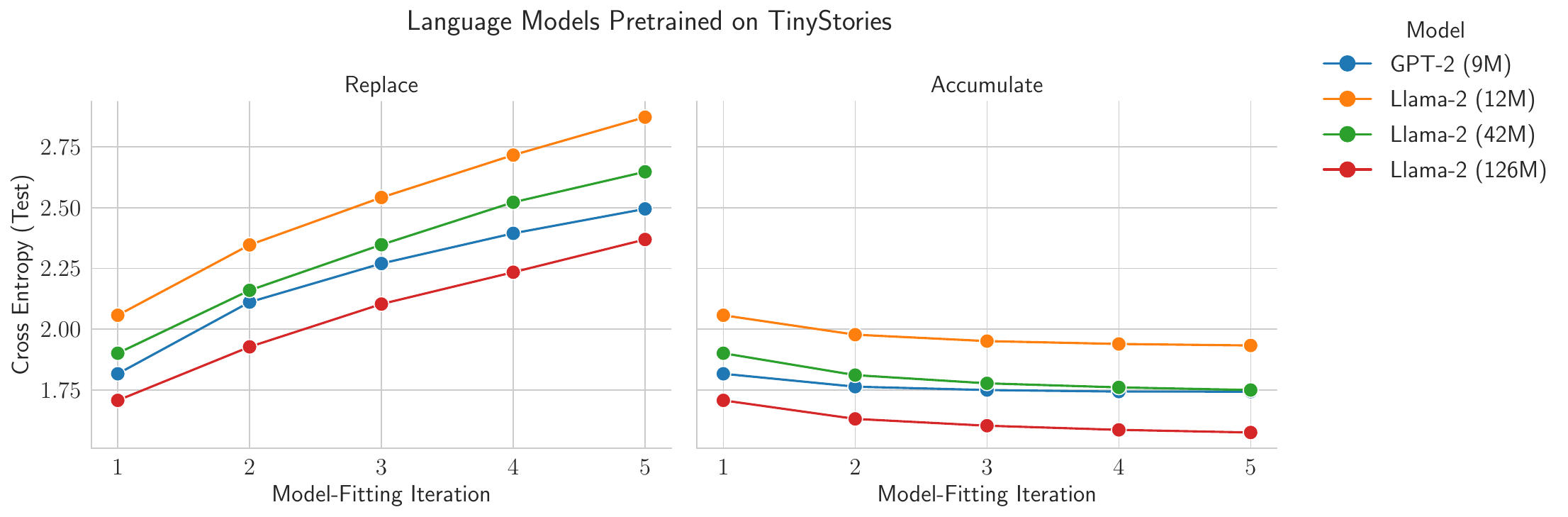}
    \caption{\textbf{Dimensions of Consideration for Model-Data Feedback Loops: Propagation of Data Over Time and Proportion of Real Data Over Time.} When data are \textit{replaced} after each model-fitting iteration (left), the proportion of real data immediately becomes zero after the first iteration, whereas when data instead \textit{accumulate} (right), the proportion of real data falls asymptotically to zero. 
    \citet{gerstgrasser2024model, kazdan2024collapsethriveperilspromises, dey2024universality} showed that replacing data over time causes the population risk to diverge,
    whereas accumulating data avoids diverging population risk. In these works, synthetic data are assumed to grow linearly over time, contributing $n$ samples per model-sampling iteration. Credit: The bottom figure is copied from  \citet{gerstgrasser2024model} with permission.}
    \label{fig:data_matters}
\end{figure*}

We believe that researchers and policy makers interested in potential societal implications of model collapse should focus on research that adhere to these conditions as faithfully as possible (with the obvious caveat that computational budgets limit research). 

%% file: 04_assessing_results.tex
\section{Key Dimensions of Consideration for Model-Data Feedback Loops}

\subsection{Propagation of Data Over Time}

Early work that sounded the alarm about model collapse \citep{martinez2023towards, alemohammad2023self, bohacek2023nepotistically, shumailov2023curse,briesch2023large, bertrand2023stability} assumed that data propagate in a particular way: after training a model, all existing data are deleted, new data are sampled from the new model, and the next model is trained solely on this freshest synthetic data.
Subsequent authors called this the \emph{replace} paradigm \cite{gerstgrasser2024model, kazdan2024collapsethriveperilspromises, dey2024universality} because data are entirely replaced after each model-fitting iteration.
When data are replaced, researchers demonstrated multiple harmful outcomes:  variances collapse, real data tails disappear, population risk diverges and so on (Fig.~\ref{fig:data_matters} left).
This particular assumption of how data propagate over time is highly unrealistic: 
\textbf{After a model finishes training, the entire internet is not deleted, nor is the next model necessarily trained solely on its predecessor's outputs.}
Rather, a more realistic assumption is that synthetic data from each model \emph{accumulates} on the internet alongside real data and past synthetic data such that all can be used for training the next model.
To some, these differences might seem insignificant, but each produces vastly different asymptotic behavior in terms of population risk:  \citet{gerstgrasser2024model} showed empirically and \citet{kazdan2024collapsethriveperilspromises} and \citet{dey2024universality} showed mathematically that population risk diverge if data are replaced, but population risk does not diverge if data instead accumulate (Fig.~\ref{fig:data_matters} right). 

A slightly different but perhaps more realistic data propagation assumption is that real and synthetic data accumulate, but future models are trained on a downsampled proportion of the total available data \citep{kazdan2024collapsethriveperilspromises}.
This \emph{accumulate-subsample} paradigm represents a middle ground between replace and accumulate: while test loss appears to stabilize, no analytic theory has been proven in this case to date.  
Individual beliefs about which data paradigm is most reflective of reality can influence perceptions of how model collapse will unfold in the future.  
However, to our knowledge, non-population risk-based notions of model collapse have not been connected to assumptions about how data propagate, leaving an important question open.

\subsection{Proportion of Real Data Over Time}

ChatGPT alone produces $1/1000$ of all words produced by humanity each day \cite{altman2024tweet}, and as these models proliferate, over time, future generative models could produce vastly more data than humanity for training future models.
Thus, the proportion of real data on the future internet plays a crucial role in the debate over the effects of model collapse.
\citet{bertrand2023stability} claimed that the population risk will not asymptotically diverge so long as the proportion of real data remains lower-bounded above $0$ (in addition to other conditions). Relatedly, in \citet{dohmatob2024strong}'s view, \textit{any} synthetic data causes ``a critical degradation" to future models, writing model collapse ``generally persists even when mixing real and synthetic data, as long as the fraction of training data which is synthetic does not vanish" and that ``model collapse cannot generally be mitigated by simple adjustments [...] unless these strategies asymptotically remove all but a vanishing proportion of synthetic data from the training process."

Results like these draw attention to what proportion of real data is necessary to avoid collapse, which is a valid consideration.
However, correctly interpreting these results takes care.
For instance, \citet{bertrand2023stability}'s condition is \textit{sufficient}, not necessary; it should \textit{not} be understood as saying that model collapse is inevitable unless real data remains a non-zero proportion.
Moreover, contrary work by \citet{gerstgrasser2024model}, \citet{marchi2024heat}, and \citet{kazdan2024collapsethriveperilspromises} established conceptually stronger guarantees: when data \textit{accumulate}, but human data asymptotically occupy a
vanishing fraction of the internet, the population risk will likely not diverge (in some cases, with an additional requirement that the rate of AI data generation does not grow super-linearly) (Fig.~\ref{fig:data_matters}).
Relatedly, \citet{gillman2024self} also show that the proportion of real data can asymptotically approach zero using a function to ``correct" synthetic data towards the real data distribution.
Due to uncertainty about the rate of synthetic data generation, one potential solution is to sequester real training data for future use, when the internet still contains an abundance of human-generated data; however, frontier AI labs have already collected such data, meaning no additional work is required.

\subsection{Model Training Assumptions}

\begin{figure*}[t!]
    \centering
    \includegraphics[width=0.49\linewidth]{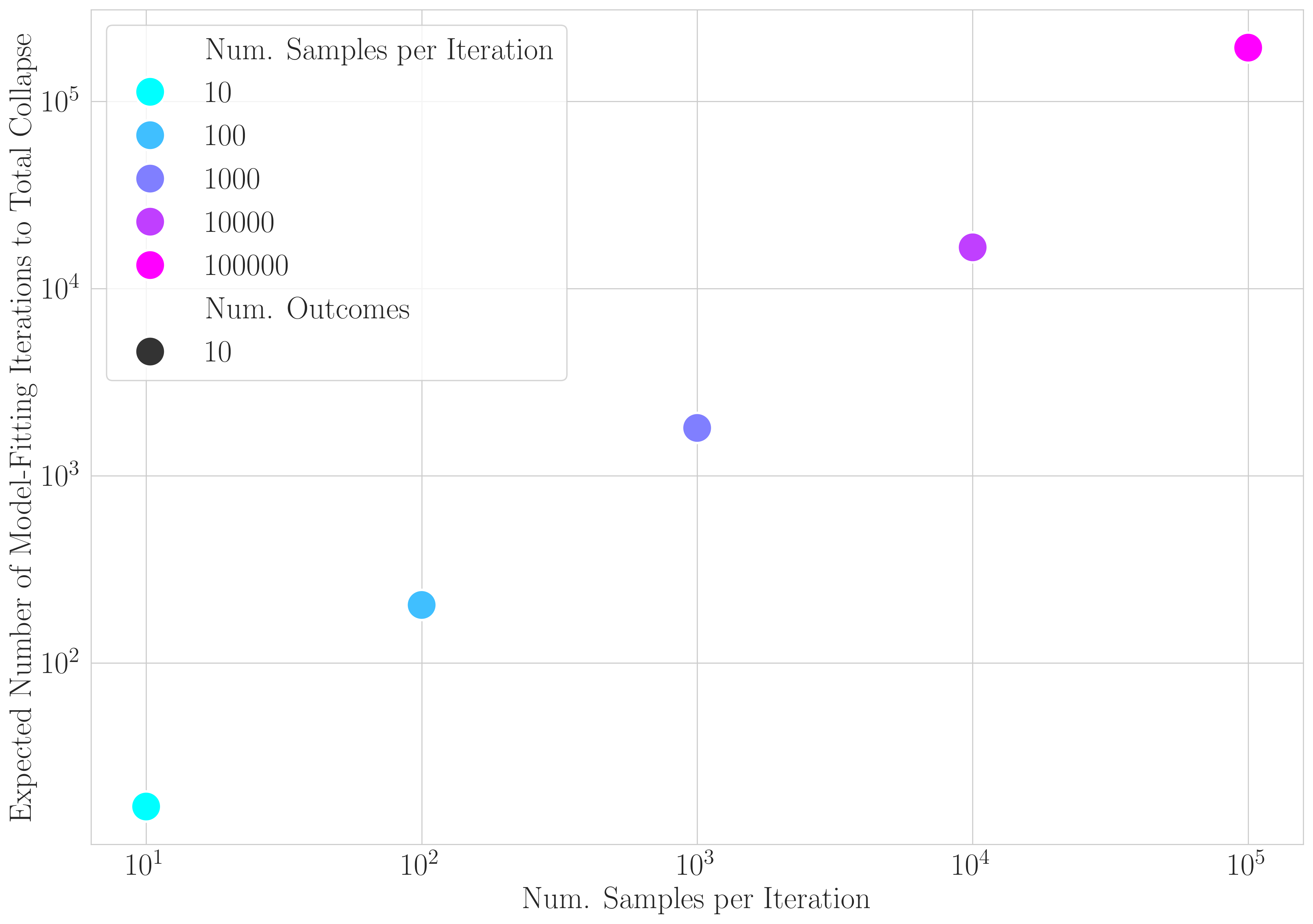}
    \includegraphics[width=0.49\linewidth]{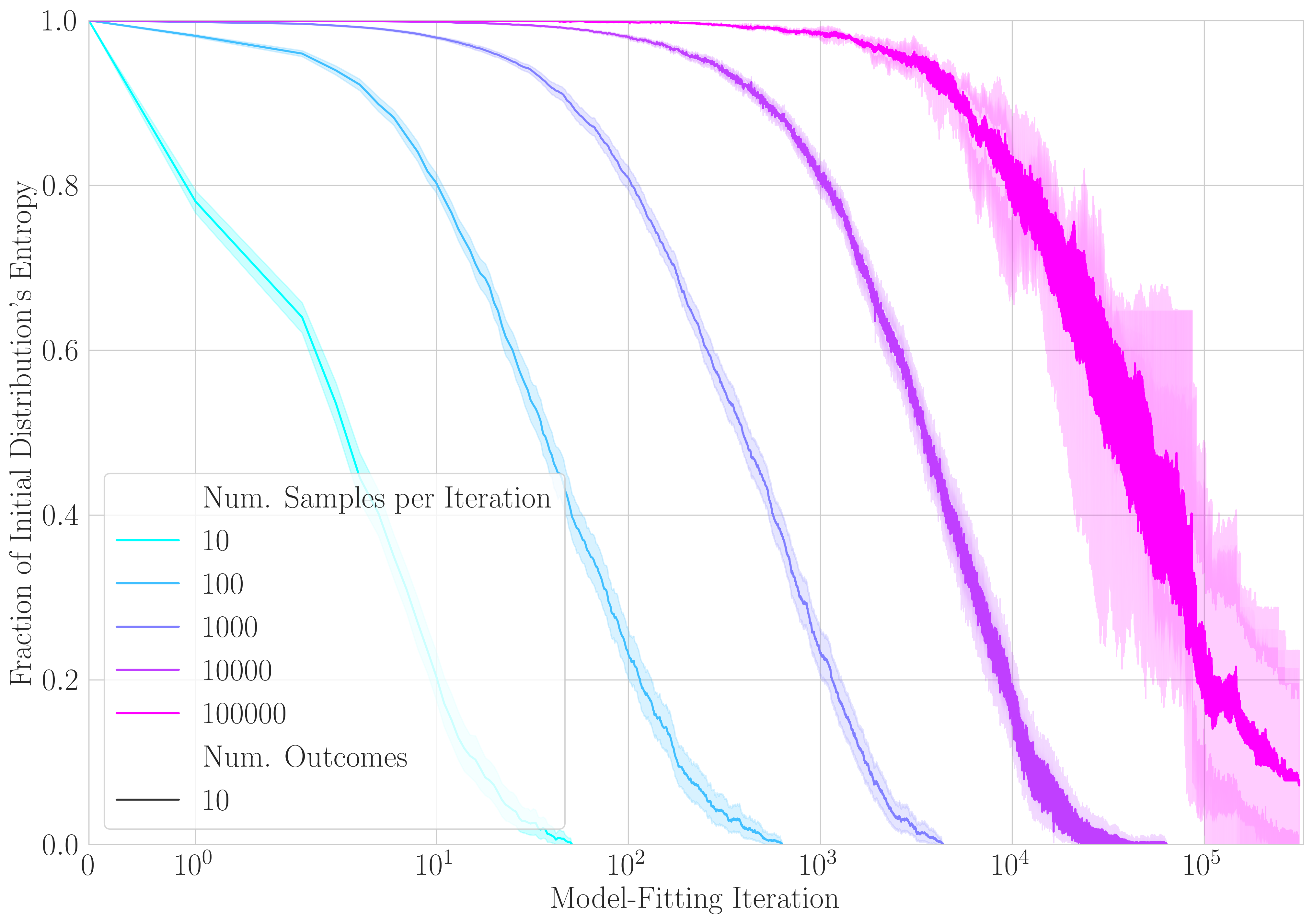}
    \caption{\textbf{Dimension of Consideration for Model-Data Feedback Loops: Timescales of Collapse.} Characterizing the timescale over which one should expect collapse is an underappreciated but crucial consideration. Focusing on the discrete model of \citet{shumailov2023curse}, the expected number of model-fitting iterations before total collapse is proportional to the number of data times the entropy of the initial data distribution (left). Taking this model at face value, this means that trillions of models can be trained before glimpsing the onset of collapse. However, total collapse is only the most extreme outcome; in this model, we additionally show how the entropy of the initial data distribution decays over time (right). Error bars are over 100 seeds (0 to 99, inclusive); for experimental details, see Sec.~\ref{sec:dims_of_consideration:subsec:timescales}.
    }
    \label{fig:timescales_matter}
\end{figure*}

While multiple works claim that models do not collapse under certain settings or propose interventions to avoid collapse, we urge the explicit declaration of strong technical assumptions that are frequently unrealistic.
As an example, \citet{bertrand2023stability} and \citet{gillman2024self} study \textit{iterative retraining}, where each model is initialized from its predecessor's parameters and optimizer state.
By making this assumption, each model is close to its predecessor; then, under an additional assumption that the first model is sufficiently high performing, since each subsequent model is close to its predecessor, model collapse can be avoided.
However, to the best of our knowledge, iterative retraining has not been used for any frontier AI model, including OpenAI's GPT-2 \citep{radford2019language}, GPT-3 \citep{brown2020language}, GPT-4 \cite{openai2023gpt4}, Anthropic's Claude 1, 2 or 3, Google's PaLM 1 \citep{chowdhery2022palmscalinglanguagemodeling}, PALM 2 \citep{anil2023palm2technicalreport} or Gemini \citep{team2024gemini}, DeepSeek's V3 \citep{deepseekai2024deepseekv3technicalreport}.
Thus, \citet{bertrand2023stability}'s sufficiency conditions for avoiding collapse are far from current practices.

For another example, \citet{zhu2024synthesizetextdatamodel} propose data editing as a collapse mitigation strategy and analyze a self-consuming linear model. Each model iteration fits $\hat{w}_{n}=X^\dagger \Tilde{Y}_n$, where $X$ are fixed Gaussian covariates, and generates synthetic data $\hat{Y}_{n+1}=X\hat{w}_n+E_{n+1}$ where $E_{n+1}$ are Gaussian errors and $\Tilde{Y}_n$ are regression targets sampled from the previous generation's training targets with edits from the prior generations synthetic data:
$$\Tilde{Y}_n^\top=M_{n-1}\hat{Y}_n+(1-M_{n-1})\Tilde{Y}_{n-1}$$
Here, $M_{k}$ is a diagonal matrix of $1$'s or $0$'s indicating whether to replace or not replace a label with a synthetic value. While \citet{zhu2024synthesizetextdatamodel} claim that this prevents model collapse, the proof of their test error bound (Theorem 2) assumes that $\|M_i\|=\eta\|M_{i-1}\|$ for some constant $\eta\in (0,1)$, meaning that the number of edits decreases by at least some fixed proportion each generation. This geometric decay guarantees that the total number of edits at any given generation is finite, and their proof fails without this. However, training GPT-2 on the Natural-Instructions dataset \citep{mishra2021cross} yields only a slight decline in edit percentage \citep{zhu2024synthesizetextdatamodel}. Moreover, if the number of edits is finite, each generation trains on mostly real data. 

\subsection{Timescales of Collapse}
\label{sec:dims_of_consideration:subsec:timescales}

Another key dimension of consideration is the timescale of model deterioration, which is often omitted. For example, \citet{shumailov2023curse} introduced a simple theoretical setting for studying model collapse: a discrete distribution (picture a histogram) with $N$ outcomes (atoms):
\begin{equation*}
    p^{(0)}(x) = \sum_{n=1}^N w_n \delta_{x_n}(x),
\end{equation*}
where $\sum_n w_n = 1$. If one sequentially draws $D$ data from this distribution and computes a new distribution based on the empirical proportions, then this process forms a Markov chain with $N$ absorbing states, each corresponding to a totally collapsed distribution, i.e., a distribution comprised of exactly one outcome. Consequently, one can use standard results from \href{https://en.wikipedia.org/wiki/Absorbing_Markov_chain}{absorbing Markov chains} to show that this simple process \textit{must} collapse.

However, one can go beyond a guarantee of collapse and ask: \textit{how many model-fitting iterations can we survive before total collapse consumes us?}
Again using standard results \citep{brydges2009wrightfisher}, the expected number of model-fitting iterations before total collapse is:
\begin{equation*}
    \mathbb{E}[\text{Model Iterations Till Total Collapse}] \propto D \, H[p^{(0)}],
\end{equation*}
where $H[\cdot]$ is the Shannon entropy. For intuition, if our starting distribution is uniform, the entropy is $\log(N)$ and thus the expected number of model-fitting iterations before total collapse is $D \log(N)$.
We confirmed this claim using numerical simulations starting from uniform distributions (Fig.~\ref{fig:timescales_matter} Left).
We also numerically simulated how quickly the entropy of $\hat{p}^{(t)}$ falls relative to the entropy of $p^{(0)}(x)$ to provide a description of the process more nuanced than just total collapse (Fig.~\ref{fig:timescales_matter} Right), since one might be interested in how quickly tail information is lost, not just how quickly total collapse arrives.

While this mathematical model is simple, for the sake of argument, if we take this model at face value, we realize total model collapse poses virtually no present threat. This is because the number of real public text data alone is on the order a quadrillion tokens \citep{villalobos2024rundatalimitsllm}, and text, image, video and agentic data are high dimensional with large entropy, meaning total collapse occurs so imperceptibly slowly that humanity could train trillions of models before noticing the onset of collapse.
However, this highlights a more general point: characterizing the timescale over which one should expect model collapse to occur is an underappreciated but crucial consideration for describing the model collapse profile.

Several authors do characterize timescales.
\citet{suresh2024ratemodelcollapserecursive} give an exact formulation of model collapse rate in the fundamental setting of recursive maximum likelihood estimation for discrete distributions and Gaussian mixtures; they find that for $\mathrm{Bern}(\mu)$, $\mathrm{Pois}(\lambda)$, and Gaussian mixtures with shared variance $\sigma^2$ distributions, the parameters $\mu$, $\lambda$, and $\sigma$ collapse to $0$ exponentially in the number of generations.
While \citet{seddik2024bad} previously controlled the total collapse probability in both the fully synthetic and partially-synthetic recursive training regimes, \citet{suresh2024ratemodelcollapserecursive} further control the number of unique symbols after $k$ generations in the fully synthetic regime. 
Lastly, \citet{kazdan2024collapsethriveperilspromises} note, in the context of kernel density estimators with fixed bandwidths, that the negative log likelihood does diverge asymptotically, although ``this occurs at a rate so glacial that it doesn’t pose a practical concern."

%% file: 05_position.tex
\section{Is Model Collapse a Threat?}
\label{sec:editorial}

In conclusion, is model collapse a threat? In our view, model collapse is a multifaceted phenomenon, and a single answer is not possible. By taking a  realism-weighted average of different papers' results, we synthesize our own forecast of model collapse under the different definitions:

\textbf{Population risk will not increase catastrophically or diverge asymptotically.} Given the increase in pretraining dataset size and quality, we argue that models training on accumulating synthetic data alongside real data will not suffer from catastrophic population risk increase or diverging population risk. The jury is still out on whether the proportion of real data relative to available data will approach zero.

\textbf{Real tail data and modes will be lost, but how many and how quickly is unclear.} Loss of diversity is a real issue, with disproportionate harms oftentimes born by subgroups. It is unclear how much of the tail we will lose or which of the real data modes will become entangled, and how synthetic data affect such changes that already occur naturally. We strongly encourage more research regarding coverage and mode collapse prevention strategies for realistic settings, building on prior work such as \citet{hashimoto2018fairness,ensign2018runaway,taori2023data}.

\textbf{Scaling laws may change with the introduction of synthetic data.} The precise nature of these changes under realistic conditions remains to be determined. Synthetic data could potentially remove what some researchers describe as a data bottleneck, but this benefit might come at the cost of altered scaling law parameters. We encourage further research into how synthetic data affects scaling behaviors to better characterize likely future outcomes.

We emphasize that while real threats do exist, the popular perception that synthetic data on the internet will render future frontier AI models pretrained on web-scale data useless is likely unrealistic since such failures appear in conditions that do not faithfully match what is actually done in practice.
Subtle degradations in data distributions might still insidiously occur, such as loss of real tail data, and future work should aim to explore what can be used to counter such outcomes.

\section{Alternative Views}
One may feel the conditions we identify in Section~\ref{sec:realistic_assumptions} are inaccurate, disproportionately emphasized, likely to change, or ignorant of important settings other than pre-training. One might also believe that the identified model collapse definitions in Section~\ref{sec:multiple_definitions} do not fully encompass the literature or are inaccurately applied in Figure~\ref{fig:definitions_meta_analysis}. Finally, a concerned bystander could argue that the benefits of being over-cautious outweigh the costs: if society fails to anticipate model collapse by allowing wanton generation of poor-quality synthetic data, then the internet could become flooded with low-quality samples that preclude future progress.
These are valid points, and we look forward to engaging with researchers and policymakers to better identify what matters to them and what the future looks like in those directions.

%% file: 99_appendix.tex
\appendix
\onecolumn

\section{A Case Study of Disagreement Between Papers} \label{app:sec:case_study}

The wide variety of non-equivalent definitions for model collapse often creates apparent contradictions between papers claiming to study the same phenomenon. A representative example presents itself in the study of model collapse for linear regression. In this data setting, one begins with a dataset $(X, y)$ where we assume that \[y\sim X\beta + \epsilon, \space \epsilon \sim \mathcal{N}(0, \sigma\cdot I).\] In the first iteration, one computes \[\hat{\beta}^{(1)} = \left(X^TX\right)^{-1}X^T y, \] and uses the fit parameter to generate new data $(X, y^{(1)})$ with \[y^{(1)} = X\hat{\beta}^{(1)}  + \epsilon_1, \space \epsilon_1\sim \mathcal{N}(0, \sigma\cdot I).\]  One can then fit successive model iterations using an \emph{accumulate} paradigm, in which the data for the $n$th model fitting takes the form
\begin{align*}
\left( \left[y,y^{(1)},y^{(2)},\dots,y^{(n-1)}\right]^\top, \left[X,X,\dots,X\right]^\top\right)
\end{align*}
One can also fit the $n$th model iteration using a \emph{replace} paradigm, in which $\hat{\beta}^{(n)}$ is computed using only the data $(X, y^{(n-1)})$.  
As proven by \citet{gerstgrasser2024model}, in the accumulate paradigm, the ratio \[\frac{\mathbb{E}\left[\lVert \hat{\beta}^{(n)}X_{\textrm{test}} - y_{\textrm{test}}\rVert^2\right]}{\mathbb{E}\left[\lVert \hat{\beta}^{(1)}X_{\textrm{test}} - y_{\textrm{test}}\rVert^2\right]}\] monotonically increases before converging to $\pi^2/6$. Citing Definitions~\ref{def:diverge} and \ref{def:var}, \citet{gerstgrasser2024model} correctly asserted that model collapse does not occur.  However, if one instead defines model collapse by Definition~\ref{def:eps}, this scenario does exhibit collapse.  

To complicate the story, \citet{dohmatob2024model} studied the same model under the replace paradigm. 
 Under the replace paradigm, Definition~\ref{def:diverge} suggests that model collapse occurs since the asymptotic risk diverges, while Definition \ref{def:var} implies that model collapse does not occur, since the replace scenario does not exhibit vanishing variance. Table~\ref{tab:def-contradict-accum-v-repl} shows how incompatible definitions lead to confusion.

\begin{table}[b!]
\caption{Model collapse occurs under some definitions, but does not occur under others for the regression setting described in Appendix \ref{app:sec:case_study}.}
\label{tab:def-contradict-accum-v-repl}
\centering
\begin{tabular}{lcr}  
\toprule
\textbf{Definition} & \textbf{Accumulate} & \textbf{Replace} \\
\midrule
Def. \ref{def:catastrophic} & \xmark & \cmark \\
Def. \ref{def:eps} & \cmark & \cmark \\
Def. \ref{def:diverge} & \xmark & \cmark \\
Def. \ref{def:var} & \xmark & \xmark \\
Def. \ref{def:scaling} & \cmark & \cmark \\
Def. \ref{def:modes} & \xmark & \xmark \\
Def. \ref{def:tails} & \xmark & \xmark \\
Def. \ref{def:hallucinate} & \xmark & \xmark \\
\bottomrule
\end{tabular}
\end{table}